\pdfoutput=1

\documentclass[11pt]{article}

\usepackage[]{acl}

\usepackage{times}
\usepackage{latexsym}
\usepackage{multirow}
\usepackage[T1]{fontenc}

\usepackage[utf8]{inputenc}

\usepackage{microtype}

\usepackage{inconsolata}

\usepackage{xspace}
\newcommand{\tsc}[1]{\textsc{#1}\xspace}
\newcommand{\ttt}[1]{\texttt{#1}}
\newcommand{\tbf}[1]{\textbf{#1}}
\newcommand{\tss}[1]{\textsuperscript{#1}}
\newcommand{\tn}[1]{\textnormal{#1}}

\newcommand{\corpus}{\tsc{Common-ToM}}
\newcommand{\system}{ReCoG}

\usepackage{enumitem}
\setlist[enumerate]{
    topsep=0pt,itemsep=-0.5ex,
    partopsep=0.5ex,parsep=0.5ex
}
\setlist[itemize]{
    topsep=0pt,itemsep=-0.5ex,
    partopsep=0.5ex,parsep=0.5ex,
}

\usepackage{booktabs}
\usepackage{colortbl}

\usepackage{mdframed}

\usepackage{listings}
\lstdefinestyle{minted}{
    basicstyle=\footnotesize\ttfamily\linespread{4},
    breaklines=true,
    columns=flexible,
    commentstyle=\color[rgb]{0.127,0.427,0.514}\ttfamily\itshape,
    identifierstyle=\color{black},
    inputencoding=latin1,
    keywordstyle=\color[HTML]{228B22}\bfseries,
    ndkeywordstyle=\color[HTML]{228B22}\bfseries,
    prebreak=\raisebox{0ex}[0ex][0ex]{\ensuremath{\hookleftarrow}},
    showstringspaces=true,
    stringstyle=\color[rgb]{0.639,0.082,0.082}\ttfamily,
    upquote=true
}
\surroundwithmdframed[
    backgroundcolor=gray!10, linecolor=gray!60,
    innertopmargin=0, innerbottommargin=0
]{lstlisting}

\usepackage{amssymb}

\usepackage{arydshln}

\newcommand{\cparagraph}[1]{
\paragraph{#1}
}

\usepackage{todonotes}

\title{Views Are My Own, but Also Yours: \\ Benchmarking Theory of Mind Using Common Ground}

\author{
    Adil Soubki\tss{\tn{$\blacklozenge\spadesuit\dagger$}}
    John Murzaku\tss{\tn{$\blacklozenge\spadesuit\dagger$}}
    Arash Yousefi Jordehi\tss{\tn{$\blacklozenge\ddagger$}}
    Peter Zeng\tss{\tn{$\blacklozenge\spadesuit\dagger$}} \\ \bfseries
    Magdalena Markowska\tss{\tn{$\clubsuit\spadesuit\dagger$}}
    Seyed Abolghasem Mirroshandel\tss{\tn{$\blacklozenge\ddagger$}}
    Owen Rambow\tss{\tn{$\clubsuit\spadesuit\dagger$}} \\
    \tss{$\blacklozenge$}Department of Computer Science
    \tss{$\clubsuit$}Department of Linguistics
    \tss{$\spadesuit$}Institute for Advanced Computational Science \\
    \tss{$\dagger$}Stony Brook University
    \tss{$\ddagger$}University of Guilan
}

\begin{document}
\maketitle

\begin{abstract}
Evaluating the theory of mind (ToM) capabilities of language models (LMs) has recently received a great deal of attention. However, many existing benchmarks rely on synthetic data, which risks misaligning the resulting experiments with human behavior. We introduce the first ToM dataset based on naturally occurring spoken dialogs, \corpus, and show
that LMs struggle to demonstrate ToM. We then show that integrating a simple, explicit representation of beliefs improves LM performance on \corpus.
\end{abstract}

\section{Introduction}
In cognitive science, theory of mind (ToM) refers broadly to the capacity to understand the mental states of others (e.g. beliefs, desires, emotions) even, crucially, when they differ from your own \citep{premack-woodruff-1978-does}. Successful human conversation is possible only because participants model each others' cognitive states (i.e., ToM) and plan utterances based on their intended audience \citep{clark-1996-using,brennan-clark-1996-conceptual,bender-gebru-etal-2021-dangers}. As a result, ToM has received increasing attention from the NLP community seeking to evaluate the capabilities of language models (LMs) on tasks inspired by the psychological literature \citep{sileo-lernould-2023-mindgames,ullman-2023-large}. 

In this paper, we introduce \corpus -- a question answering benchmark based on naturally-occurring, spoken dialogs in English.\footnote{\url{https://github.com/cogstates/common-tom}}
\corpus uses the notion of \tbf{common ground} (CG) for evaluating \tbf{ToM}.  The CG \citep{wilkes-gibbs/clark:1992,stalnaker-2002-common} is a set of beliefs mutually shared by all participants in a conversation. Other benchmarks have developed ToM questions based on the Sally-Anne test \citep{wimmer-perner-1983-beliefs,baron-cohen-etal-1985-does},
which checks an understanding of information accessibility to determine beliefs. 
We observe that when the CG is mismatched between the discourse participants, such as at the time of a question or during the repair of a miscommunication, 
similarly complex problems for ToM arise.
\looseness=-1


Our main contributions are: (1) arguing that using synthesized data in evaluating the ToM ability of LMs is not conclusive; (2) releasing a corpus for benchmarking ToM based on naturally occurring spoken conversations; (3) showing that LLMs struggle with our benchmark and a simple explicit architecture performs better.

The paper is organized as follows. First, we review some of the relevant literature (\S~\ref{sec:related-work}). We then describe the framework and methods used in creating \corpus (\S~\ref{sec:corpus-creation}). This is followed by experiments and results in Section~\ref{sec:experiments}: human performance, using zero-shot and fine-tuned LMs, and using our approach for building a system that explicitly represents beliefs (\S~\ref{sec:full-rep}). Finally, we discuss our key takeaways and conclusions (\S~\ref{sec:conclusion}).

\section{Related Work}
\label{sec:related-work}
While quite a few corpora annotate author belief \citep{de-marneffe-etal-2019-commitmentbank,sauri-pustejovsky-2009-factbank}, there are relatively few corpora annotated explicitly for CG. 
\citet{horton-gerrig-2016-revisiting,soubki-etal-2022-kojak} focus on limited aspects of CG.  In this paper, we use the comprehensive CG corpus we presented in \citep{markowska-etal-2023-finding}, which we discuss in Section~\ref{sec:corpus-creation}. 

Many ToM benchmarks have been created recently. \citet{nematzadeh-etal-2018-evaluating} produce a question answering corpus (ToM-bAbi) derived from template-generated stories inspired by the Sally-Anne test (an agent's cognitive state depends on whether they witness a specific action). \citet{le-etal-2019-revisiting} note that such formulaic data results in a flawed evaluation, especially when using supervised methods, and produce their own templatic corpus (ToMi) which introduces more noise such as distractor sentences and reorderings. Despite these improvements, ToMi has been shown to be prone to the same issues as ToM-bAbi \citep{sclar-etal-2023-minding}. \citet{kim-etal-2023-fantom} 
take this even further and prompt LMs to produce dialogs following a similar form. The stories all follow the general structure of the Sally-Ann stories.
Corresponding questions then probe various character's beliefs (true and false) of both first and second order. 

Though useful, these benchmarks are prone to surface-level cues and spurious correlations which have been exploited by LMs to display \emph{illusory ToM} \citep{kosinski-2023-theory}. Some work has been done to produce tools based on human-generated text. 
\citet{bara-etal-2021-mindcraft}, like us, exploit the relationship between CG and ToM. They design an experimental setup that records written dialogs of players exchanging information in MineCraft.
More recently, \citet{ma-etal-2023-holistic} combine the aforementioned datasets (and more) to create a composite benchmark. We build on this work conceptually by producing a new dataset which, in contrast with the work of  \citet{bara-etal-2021-mindcraft}, is based on dialogs which are collected independently of our interest in ToM; the dialogs are spontaneous and not guided by any experimental setting; and they are spoken dialogs. We note that co-presence (including telepresence) has been identified as an important (though not required) part of human ToM \citep{galati-brennan-2021-retained}, and our corpus allows the study of ToM under this common condition using naturalistic data. \looseness=-1

\begin{table*}[htb]
\ttfamily \small
\centering
\resizebox{1.0\textwidth}{!}{
\begin{tabular}{lcccc}
\hline
\bfseries Transcript \hspace{31.1em} [Event: B is now not smoking] &
\tbf{\tsc{Bel$_\textrm{A}$}} &
\tbf{\tsc{Bel$_\textrm{B}$}} &
\tbf{\tsc{CG$_\textrm{A}$}} &
\tbf{\tsc{CG$_\textrm{B}$}} \\
\hline
\rowcolor{red!10}
114 | B: Small world \%huh? & NB & CT- & NA & NA \\ \rowcolor{red!10}
115 | A: You’re & NB & CT- & NA & NA \\ \rowcolor{red!10}
116 | A: yeah. And now are you not smoking? & NB & CT- & NA & NA \\
\hline
\rowcolor{violet!15}
117 | B: No. I am smoking. & CT- & CT- & RT & RT \\ \rowcolor{violet!15}
118 | A: yeah. Well I did a few when I’ve been back too. & CT- & CT- & RT & RT \\ \rowcolor{violet!15}
119 | B: yeah. I’ve been smoking big time. It’s been a rough couple months. & CT- & CT- & RT & RT \\ 
\hline
\bfseries Question & \multicolumn{2}{c}{\tbf{Before}} & \multicolumn{2}{c}{\tbf{After}} \\
\hline
\cellcolor{gray!10} 1st Order Question: Does A believe it is certainly not true that B is now not smoking? &
\multicolumn{2}{c}{\cellcolor{red!10} No} &
\multicolumn{2}{c}{\cellcolor{violet!15} Yes} \\
\cellcolor{gray!10} 2nd Order Question: Does B believe that A believes it is certainly not true that B is now not smoking? &
\multicolumn{2}{c}{\cellcolor{red!10} No} &
\multicolumn{2}{c}{\cellcolor{violet!15} Yes} \\
\hline
\end{tabular}
}
\caption{Example dialog extract with belief and common ground annotations from the corpus for the event ``B is now not smoking''. Some of our derived ToM questions are below.}
\label{tab:example-questions}
\end{table*}

\section{\corpus}
\label{sec:corpus-creation}
\label{sec:answer-heuristics}



\cparagraph{Framework}
In philosophy, CG is sometimes treated as the mutual beliefs between two agents \cite{stalnaker-2002-common}, separate from their cognitive states, while in cognitive science it is common 
to model CG as the belief of an agent about what they and another agent mutually believe \citep{brown-schmidt-duff-2016-memory}.
We adopt the latter definition as it is allows CG to represent scenarios like false belief, which would be impossible under the former. This also makes the relationship between CG and ToM explicit: If I believe a proposition is in the CG with you, it means by definition that I believe that you also believe it is in CG – thus, it entails a ToM assumption.

We also follow cognitive literature in assuming that humans do not have a full representation of their interlocutor’s cognitive state at all times, but instead can make necessary inferences when needed \citep{horton-brennan-2016-role}. Specifically, CG does not mean that all consequences of CG are present at all times, and performing the inferences from CG (high-degree knowledge questions, for example) takes time and may be errorful. Our experiments with human annotators (\S~\ref{sec:experiments}) confirm this assumption, but also show that humans are much better than zero-shot LMs because they are in fact modeling CG, and can answer the higher-order belief questions as needed from their models of CG. 
\citet{farkas-bruce-2010-reacting,eckardt-2016-questions} propose deterministic heuristic algorithms for inferring CG from text. All this prior work motivates our neuro-symbolic approach (\S~\ref{sec:full-rep}).

\cparagraph{Base Corpus}
In our previous annotation work \citep{markowska-etal-2023-finding}, we use a corpus of 1,710 turns of dyadic dialog across four conversations from CALLHOME, and annotate them for CG. We extract events (i.e., propositions) evoked by each turn and label the beliefs of each discourse participant (DP) towards those events as certainly true (\tbf{CT+}), possibly true (\tbf{PS}), certainly not true (\tbf{CT-}), or we label the DP as not having a belief (\tbf{NB}). Then 
we determine for each DP if the speaker has just added the event to their view of the CG (\tbf{JA}), already has the event in the CG (\tbf{IN}), or rejects it from the CG (\tbf{RT}). 
As we do not annotate CG for events before they are introduced in \citep{markowska-etal-2023-finding}, we use \tbf{NA} to indicate this in this study.  
Note that the annotation assumes that the CG is not actually shared, but rather independently hypothesized by each DP.
\footnote{We refer the reader to Appendix~\ref{sec:detailed-corpus-stats} and \citep{markowska-etal-2023-finding} for additional details regarding the corpus.}\looseness=-1


\cparagraph{Query Creation}
To motivate the query creation process, consider the examples in Table~\ref{tab:example-questions}. The proposition under question is ``B is now not smoking''. The annotations from the CG corpus for this proposition throughout discourse time (which we simply measure in turns) are shown above, and resulting first order (A/B believes) and second order (A/B believes that B/A believes) questions are shown below.



Queries are created in three steps. For each proposition extracted from the dialog we (1) identify where the proposition is introduced and every point where one of the agent's belief or CG about the proposition changes (``utterance of interest''). In this case that is at times 116 (introduction) and 117 (change). We then (2) generate queries, up to third order, for those points in time.
{\addtolength\leftmargini{-0.27in}
\begin{quote}
\small
\ttfamily
\noindent
At the time indicated, is it the case that (((A/B believes that)$^1$ B/A believes that)$^2$ A/B believes that)$^3$ it is \{certainty\} true that \{proposition\}?
\end{quote}}
\noindent
For each proposition and for each point in the conversation selected for it, we vary \ttt{certainty} (options: certainly, possibly, certainly not).  We then generate the first, second, and third-order belief questions as shown in the template above by the superscripts.
We omit queries which ask about self-belief (e.g. A believes A believes). 

Therefore there are two query templates per order and we have three orders, making six templates. Each of these templates is instantiated with three possible \ttt{certainty} values, resulting in 18 total queries per proposition and selected point in the conversation. By asking multiple questions regarding a single proposition throughout time we can evaluate the consistency of model responses. As the majority of propositions are, uninterestingly, labeled CT+ and JA for both speakers, the final corpus samples just 10\% of these instances.

Finally, (3) we determine the answers to the generated queries using a set of rules. For first order questions this is straightforward and we simply check that the belief annotation matches the question, with one caveat. If A believes proposition $p$ is certainly true and the query asks if A believes $p$ is possibly true, we consider the answer to be yes. For higher order queries we must use the CG annotations.
To illustrate, consider the 2nd order question whether A believes that B believes some proposition $p$.
(1) If the \ttt{certainty} is positive polarity (i.e., certainly or possibly) and A believes the proposition to be CG (i.e., JA or IN), then resolve the answer just as in the first order case described above. (2) Otherwise, if the \ttt{certainty} is negative polarity (i.e. certainly not) and A is aware it was rejected from CG (i.e., RT) and the \ttt{certainty} of the query matches the \ttt{certainty} of B, then label the question yes. (3) For all other cases, label the answer no. The whole set of heuristics is in Appendix~\ref{sec:appendix-code}.\looseness=-1

\begin{table}[t]
\centering
\begin{tabular}{|c|c|c|}
\hline
\textbf{Split}         & \textbf{Answer} & \textbf{Count} \\ \hline
\multirow{2}{*}{\textbf{Train}} & No             & 2899           \\ \cline{2-3} 
                       & Yes            & 2371           \\ \hline \hline
\multirow{2}{*}{\textbf{Test}}  & No             & 1139           \\ \cline{2-3} 
                       & Yes            & 965            \\ \hline
\end{tabular}
\caption{Number of questions in \corpus with yes and no answers broken down by split.}
\label{tab:train-test-split}
\end{table}

\cparagraph{Corpus Statistics}
This results in 7,374 queries to probe for yes/no answers regarding the beliefs of speakers from CALLHOME. 
There are a roughly equal number 
of first, second, and third order queries.
The corpus is partitioned using the same splits as was done in \citep{markowska-etal-2023-finding} -- three conversations for training, and a held out fourth conversation for testing. The counts by query answer are shown in Table~\ref{tab:train-test-split}.
Additional information regarding the data and splits is in Appendix~\ref{sec:detailed-corpus-stats}.\looseness=-1

\section{Experiments}
\label{sec:experiments}

\begin{table*}[!htb]
\centering
\resizebox{0.85\textwidth}{!}{
\begin{tabular}{|l|c|c|c|c|}
\hline
\textbf{\begin{tabular}[c]{@{}c@{}}Model\end{tabular}} & \textbf{\begin{tabular}[c]{@{}c@{}}Total \end{tabular}} & \textbf{\begin{tabular}[c]{@{}c@{}}First Order \end{tabular}} & \textbf{\begin{tabular}[c]{@{}c@{}}Second Order \end{tabular}} & \textbf{\begin{tabular}[c]{@{}c@{}}Third Order \end{tabular}} \\ \hline\hline

Random Baseline & 50.4 & 50.3 & 50.5 & 50.4 \\
\hline
\hline
gpt-3.5-turbo-0613 (Zero-Shot) & 57.0 & 60.7 & 57.7 & 53.0 \\
\hline
gpt-4-0613 (Zero-Shot) & 63.4 & 65.5 & 62.5 & 62.1 \\ \hline

Mistral-7B-Instruct (Zero-Shot) & 60.6 & 63.3 & 60.5 & 58.0 \\
\hline
\hline

Mistral-7B (Fine-Tune) & 64.0 & 64.8 & 63.9 & 63.2 \\
\hline
\hline
\system & \textbf{71.0} & \textbf{70.4} & \textbf{71.3} & \textbf{71.2} \\

\hline\hline
\textit{Human Performance} & \textbf{\textit{80.0}} & \textbf{\textit{85.0}} & \textbf{\textit{80.0}} & \textbf{\textit{75.0}} \\
\hline
\end{tabular}}
\caption{\label{table:results}
        Experimental results of \corpus on different models. We report total accuracy (Total) and per-order accuracy. We bold our best results. We compare all results to the random baseline and to our Human Performance baseline.
    }
\end{table*}

\subsection{Setup} 
\cparagraph{Data}
For fine-tuning experiments we use the train and test splits for \corpus as described in Section~\ref{sec:corpus-creation}.
When prompting, we evaluate only on the test conversation to maintain comparability. For all experiments we do not do any hyperparameter search or hyperparameter tuning.

\cparagraph{Random Performance}
We perform a random baseline by randomly selecting ``Yes'' or ``No'' answers with their probabilities proportional to their respective frequencies in the training set.

\cparagraph{Human Performance}
We randomly sample sixty queries from our test set with twenty questions from each order. Annotators are two graduate students instructed to answer the queries as best they can. We report accuracy for these sixty questions.

\subsection{Zero-Shot}
We perform zero-shot experiments using gpt-3.5-turbo-0613 (GPT-3.5) \citep{OpenAI2022ChatGPT}, gpt-4-0613 (GPT-4) \citep{openai2023gpt}, and Mistral-7B-Instruct \citep{jiang2023mistral}. 

We format our prompt by starting with an instruction (which we keep the same for all prompts), a dialog containing the utterance of interest with five previous and future utterances, and the question. 
For our GPT-3.5 and GPT-4 experiments, we use the default API hyperparameters. We provide further experimental details, prompts, and hyperparameters, as well as details on unsuccessful chain-of-thought experiments, in Appendix~\ref{sec:appendix-experimental-details}.

\subsection{Fine-Tuning} We fine-tune Mistral-7B using LoRA \citep{hu2021lora} on the train split of \corpus. We follow a similar format to our zero-shot experiments where we use a prompt, a dialog containing the utterance of interest with five previous and future utterances, and the question with its respective yes or no answer. At test time, we present our model with a dialog-question pair, and generate the yes or no answer. We provide further experimental details and hyperparameters in Appendix~\ref{sec:appendix-experimental-details}.

\subsection{Our System: \system}
\label{sec:full-rep}

Our Full-Representation approach (\system) creates an explicit \tbf{re}presentation of the \tbf{cog}nitive states of the discourse participants, and then uses rules to answer the questions.  We closely follow our previous work in \citep{markowska-etal-2023-finding} in fine-tuning FLAN-T5 and we similarly use a speaker-based window. Specifically, our model receives as input all utterances preceding and/or following the target event as input until it encounters an utterance from the other speaker. 

\noindent
Our system has three  parts: belief prediction, CG prediction, and yes/no question answering. 

\cparagraph{Belief Prediction}
We first predict the beliefs of the discourse participants towards a target event at each utterance in the dialog. Our input is as follows:\looseness=-1
\vspace{0.5em}
\begin{lstlisting}[breaklines=false]
"Preceding Context": {Preceding Utterances}
"Target Event": {Target Event}
"Following Context": {Following Utterances}
\end{lstlisting}
We treat the belief prediction task as a text generation task where given the above input of context and a target event, we generate the belief label. 

\cparagraph{CG Prediction}
To predict CG, we use the heuristics-based approach we used in  our previous work \citep{markowska-etal-2023-finding}. This rule-based approach maps a label for speaker and hearer belief to a CG label.\footnote{See Appendix~\ref{sec:appendix-heuristics} for details on this algorithm.}

\cparagraph{Yes/No Question Answering}
Now that we have the belief and CG for both discourse participants, we apply the same heuristics we used to determine the query answers for the gold-annotated case, as described in Section~\ref{sec:answer-heuristics}.

\subsection{Results}
All results are evaluated using accuracy. Specifically, we report results on total accuracy and per-order accuracy. All of these metrics help us quantify to what extent LMs capture the full mental states of others, up to third-order mental states. Results for our experiments are shown in Table~\ref{table:results}. 
All of our models perform worse than human performance, but better than the random baseline.  Humans perform best on first order beliefs, 
but performance decreases on higher orders as has been observed in similar studies \cite{valle-etal-2015-theory}.

\cparagraph{Zero-Shot}
GPT-4 performs the best on all metrics. Mistral-7B-Instruct, despite being only 7B parameters, performs better than the 175B parameter GPT-3.5 in a zero-shot setting. We notice a distinct trend: all models capture first order beliefs the best, and decrease in performance as the order increases. \looseness=-1

\cparagraph{Fine-tuning}
Compared to the best zero-shot performance from GPT-4, our results show an increase in total accuracy, second order beliefs, and third order beliefs. However, we perform slightly worse in modeling first order beliefs. 

\cparagraph{ReCoG}
ReCoG, which explicitly models beliefs and CG among agents in dialog, outperforms every other system. We see a clear boost in performance in all metrics, and more notably similar results among all orders of belief. 


\cparagraph{Discussion}
There are some interesting trends in these results. All the LM-only models see a decrease in performance as the order of the query increases, just as humans do, though this decrease is not as large as is observed for humans. This might suggest that the models are in some ways ``human-like'' in their mistakes. However, upon closer inspection we see further differences.

Since the same proposition is asked about for multiple orders of belief, we can look at how often the model gets all answers for a proposition correct. Even though fine-tuning Mistral only provides a roughly 3 point boost to overall accuracy, this consistency in answers changes dramatically with Mistral, going from getting all questions for a proposition correct 20\% of the time in the zero-shot setting to 61\% of the time when fine-tuned. Our human sample is too small to perform a similar comparison but \citet{kim-etal-2023-fantom} observed human consistency to be in the high 80s for a similar task on their synthetic benchmark.

We can also look at how correctness for first order belief correlates with correctness of higher order beliefs. Again we see fine-tuning makes a larger difference than accuracy suggests. The correlation between per-proposition 1st and 2nd order accuracy goes from $r=0.43$ for the zero-shot model to $r=0.90$ for the fine-tuned model. An even larger difference is observed between 1st and 3rd order ($r=0.20$ to $r=0.93$).

This all suggests that the zero-shot LM behavior is markedly different from the fine-tuned model, and from humans.
Even when making correct predictions, they do so using a very different pattern.



\section{Conclusion}
\label{sec:conclusion}

We present a new corpus for testing theory of mind (ToM) capabilities, \corpus.  Unlike previous ToM corpora, \corpus uses naturally occurring linguistic data and is not based on agents perceiving certain information or not.
We 
explicitly model belief and CG to capture ToM in a manner directly inspired by cognitive science literature. While LMs are rapidly evolving and scaling in size and capabilities, they still lack a conceptual understanding of beliefs and CG in dialog. 
Therefore, for an LM to interact naturally it is necessary to explicitly model belief, CG, and cognitive states.


\section*{Limitations}

While we present preliminary results on \corpus, the first ToM corpus to use natural spoken conversation, the original CG corpus is relatively small, containing only four dialogs. Furthermore, as these dialogs only contain conversations in English, our benchmark does not involve tracking ToM in other languages. While the benchmark tests ToM reasoning in a variety of ways, it is by no means comprehensive. In cognitive science, ToM is divided into cognitive (e.g. beliefs, thoughts) and affective (e.g. emotions, desires) with some evidence for fairly independent processes which operate over these two domains \cite{kalbe-etal-2010-dissociating}. We evaluate the cognitive aspect of belief and plan to tackle additional areas, particularly affective, in future work.

\section*{Ethical Considerations}

As with other work on ToM, we risk the misinterpretation that AI models may be anthromorphized as having near-human level cognition.  We stress that our work instead shows that as of now, existing models in fact do poorly at demonstrating ToM when presented with natural conversations. 
We did not require annotators for the creation of \corpus as our corpus was derived heuristically from the common ground (CG) corpus. However, our human baseline was done in-house by trained graduate students who were paid. 

\section*{Acknowledgements}
This material is based upon work supported in part by the National Science Foundation (NSF) under No. 2125295 (NRT-HDR: Detecting and Addressing Bias in Data, Humans, and Institutions); by funding from the Defense Advanced Research Projects Agency (DARPA) under the CCU (No. HR001120C0037, PR No. HR0011154158, No. HR001122C0034) and INCAS (HR01121C0186, No. HR001120C0037, and PR No. HR0011154158) programs; as well as by the Intelligence Advanced Research Projects Activity (IARPA) under the HIATUS program (contract 2022-22072200005).  Any opinions, findings and conclusions or recommendations expressed in this material are those of the author(s) and do not necessarily reflect the views of the NSF, DARPA, or IARPA.

We thank both the Institute for Advanced Computational Science and the Institute for AI-Driven Discovery and Innovation at Stony Brook for access to the computing resources needed for this work. These resources were made possible by NSF grant No. 1531492 (SeaWulf HPC cluster maintained by Research Computing and Cyberinfrastructure) and NSF grant No. 1919752 (Major Research Infrastructure program), respectively.

We thank our ARR reviewers, whose comments have contributed to improving the paper.
\bibliography{anthology,custom}

\appendix

\section{Corpus Details}
\corpus builds on \citep{markowska-etal-2023-finding}, which uses the English-only LDC CALLHOME corpus as a base.
We show counts for our train/test split in Table~\ref{tab:train-test-split}. The training split contains dialogs 4245, 4248, and 4310 while the test split contains dialog 4431.
\label{sec:detailed-corpus-stats}


For convenience, we summarize the label meanings in Table~\ref{tab:corpus-labels} and the distribution of labels for \corpus in Table~\ref{tab:common-tom-counts}. 
\begin{table}[htb]
\centering
\resizebox{1.0\linewidth}{!}{
\begin{tabular}{ll}
\hline
\bfseries Label & \bfseries Description \\
\hline
CT+ & A speaker certainly believes that $e$. \\
PS & A speaker possibly believes that $e$. \\
CT- & A speaker certainly believes that not $e$. \\
NB & A speaker expresses no belief about $e$. \\
\hline
JA & An event $e$ is mutually believed by both interlocutors \\
   & and is added to CG in the moment $e$ was uttered. \\
IN & An event $e$ has already been a part of the interlocutors’ \\
   & CGs before the time of the event. \\
RT & An event $e$ that has been presented by a speaker \\
   & has been entertained but rejected by the addressee \\
NA & An event $e$ that has no annotation for CG \\
   & because it has not yet been introduced. \\
\hline
\end{tabular}
}
\caption{Summary of annotation label meanings as described in \citep{markowska-etal-2023-finding}.}
\label{tab:corpus-labels}
\end{table}

\begin{table}[htb]
\begin{tabular}{llllr}
\toprule
\bfseries \tbf{\tsc{Bel$_\textrm{A}$}} & \bfseries \tbf{\tsc{Bel$_\textrm{B}$}} & \bfseries \tbf{\tsc{CG$_\textrm{A}$}} & \bfseries \tbf{\tsc{CG$_\textrm{B}$}} & \bfseries Count \\
\midrule
CT- & CT- & RT & RT & 1,746 \\
CT+ & CT+ & JA & JA & 1,632 \\
CT+ & CT+ & IN & IN & 1,494 \\
PS & PS & JA & JA & 1,152 \\
PS & CT+ & JA & JA & 180 \\
PS & CT+ & NA & NA & 144 \\
CT- & NB & NA & NA & 90 \\
PS & PS & NA & NA & 90 \\
NB & CT- & NA & NA & 72 \\
PS & CT- & NA & NA & 72 \\
CT+ & NB & NA & NA & 72 \\
NB & CT+ & NA & NA & 54 \\
PS & NB & NA & NA & 54 \\
CT+ & CT- & RT & RT & 54 \\
CT+ & PS & JA & JA & 54 \\
CT- & PS & NA & NA & 54 \\
CT- & PS & RT & RT & 36 \\
CT+ & PS & NA & NA & 36 \\
NB & PS & NA & NA & 36 \\
PS & CT- & RT & RT & 36 \\
PS & PS & IN & IN & 36 \\
CT+ & CT+ & NA & NA & 36 \\
CT+ & CT+ & JA & IN & 18 \\
NB & CT+ & JA & JA & 18 \\
CT- & PS & NA & JA & 18 \\
CT+ & NB & RT & RT & 18 \\
CT+ & CT- & IN & RT & 18 \\
CT- & CT+ & RT & RT & 18 \\
CT- & CT+ & NA & NA & 18 \\
PS & NB & RT & RT & 18 \\
\midrule
\multicolumn{4}{l}{\bfseries Total} & 7,374 \\
\bottomrule
\end{tabular}
\caption{Counts of belief labels included in the corpus.}
\label{tab:common-tom-counts}
\end{table}

\section{Heuristics for Common Ground Prediction using Beliefs}
\label{sec:appendix-heuristics}

Within this strategy, we've employed the following rules: consistently updating the common ground for both speakers using these straightforward heuristics.

\begin{enumerate}[label={\arabic*)}]
\item If \textbf{{\footnotesize Bel(A) = CT-}} or \textbf{{\footnotesize Bel(B) = CT-}},  then \textbf{{\footnotesize CG = RT}}. 
\item If \textbf{{\footnotesize Bel(A) = CT+}} and \textbf{{\footnotesize Bel(B) = CT+}}, \\ then \textbf{{\footnotesize CG = JA}} or \textbf{{\footnotesize CG = IN}}.
\item If \textbf{{\footnotesize Bel(A) = PS}} and \textbf{{\footnotesize Bel(B) = CT+}}, \\ then \textbf{{\footnotesize CG = JA(PS)}}  or \textbf{{\footnotesize CG = IN}}.
\item If \textbf{{\footnotesize Bel(A) = CT+}} and \textbf{{\footnotesize Bel(B) = PS}}, \\ then \textbf{{\footnotesize CG = JA(PS)}} or \textbf{{\footnotesize CG = IN}}.
\item If \textbf{{\footnotesize Bel(A) = NB}} or \textbf{{\footnotesize Bel(B) = NB}},  then \textbf{{\footnotesize CG = NULL}}.
\end{enumerate}
\noindent
Rules 2-4 under-determine whether the belief is already in the CG or newly added. In this context, the crucial task is to determine whether the target event had already been present in the common ground of the speakers (i.e., \textbf{{\footnotesize CG = IN}}) or not (i.e, \textbf{{\footnotesize CG = JA}}). 

\section{Experiment Details}
\label{sec:appendix-experimental-details}
 All experiments besides our OpenAI experiments used our employer's GPU cluster. We performed experiments on a Tesla V100-SXM2 GPU. Compute jobs typically ranged from 20 minutes for zero-shot experiments to 16 hours for fine-tuning. We do not do any hyperparameter search or hyperparameter tuning. 
 
\paragraph{Zero-Shot} For our zero-shot experiments, we use OpenAI models gpt-3.5-0613 and gpt-4-0613, and the instruction tuned Mistral-7B-Instruct \citep{jiang2023mistral}. Our OpenAI experiments use the OpenAI API. We set temperature to 1.0. To perform zero-shot experiments on Mistral-7B-Instruct, we use HuggingFace transformers \citep{wolf-etal-2020-transformers} and use the same temperature hyperparameter of 1.0. We keep all other hyperparameters as default. Our prompt template is as follows:

\vspace{0.5em}
\begin{lstlisting}[breaklines=false]
You are a cautious assistant. You carefully 
follow instructions. You are helpful and 
harmless and you follow ethical guidelines 
and promote positive behavior. Given a 
conversation, answer a yes or no question
without providing any additional 
information.

Conversation: 
{context}

Question:
{question}
\end{lstlisting}

We perform experiments using both chain-of-thought (CoT) following \citep{kim-etal-2023-fantom} and without CoT. CoT experiments are performed by adding  “let’s think step by step” after the question. Our CoT experiments performed 1.3\% worse on average than without CoT, so therefore we only report results without CoT. 

\paragraph{Fine-tuning} We fine-tune the base Mistral-7B-v0.1 model. We use LoRA \citep{hu2021lora} to make our model memory efficient with \textit{r=32} and \textit{alpha=64}. We fine-tune for 3 epochs and use a learning rate of 2e-5. 
\paragraph{ReCoG} We perform a standard fine-tuning approach of FLAN-T5-base model which has 250M parameters. We fine-tune for 12 epochs and use a learning rate of 3e-4. Our system is implemented using the PyTorch
framework and Python programming language. FLAN-T5 model is captured from the HuggingFace transformers \citep{wolf-etal-2020-transformers}. We also used Pandas data analysis library \cite{reback2020pandas}.





\section{Human Evaluation Annotators}
\label{appendix:annotators}
Our human evaluation annotation was done in-house with two domain expert graduate students who volunteered to annotate. The annotators were given instructions to answer 30 randomly sampled yes/no questions as best as they can. Annotators received ten questions of each order. Reported accuracy is computed over all 60 questions.

\section{Query Creation Heuristics}
\label{sec:appendix-code}
\begin{lstlisting}[
    language=python,
    emph={resolve_1st_order_yn_answer},
    emphstyle=\color{blue}
]
def resolve_1st_order_yn_answer(qbel, sbel):
    if qbel == sbel:
        return True
    elif qbel == "PS" and sbel == "CT+":
        return True
    else:
        return False
\end{lstlisting}
\begin{lstlisting}[
    language=python,
    emph={resolve_2nd_order_yn_answer},
    emphstyle=\color{blue}
]
def resolve_2nd_order_yn_answer(qbel, sbel1, sbel2, cg1, cg2):
    """
    This method resolves the truthiness of a second order belief question with
    the following form based on the annotations provided.

        Is it the case that {spkr1} believes that
        {spkr2} believes that it is {qbel} true that {event}?

    Args:
        qbel (str): question belief (CT-, PS, CT+).
        sbel1 (str): speaker 1 belief (CT-, PS, CT+, NB).
        sbel2 (str): speaker 2 belief (CT-, PS, CT+, NB).
        cg1 (str): speaker 1 common ground (JA, IN, RT, NA).
        cg2 (str): speaker 2 common ground (JA, IN, RT, NA).

    Returns:
        bool: True if the question is correct given the annotations.
    """
    # Case 1: The question is positive polarity.
    if qbel in ("PS", "CT+") and cg1 in ("JA", "IN") and (
        (qbel == sbel1) or ((qbel == "PS") and (sbel1 == "CT+"))
    ):
        return True
    # Case 2: The question is negative polarity.
    elif qbel == "CT-" and cg1 == "RT" and qbel == sbel2: # XXX: sbel2 here!!!
        return True
    return False
\end{lstlisting}
\begin{lstlisting}[
    language=python,
    emph={resolve_3rd_order_yn_answer},
    emphstyle=\color{blue}
]
def resolve_3rd_order_yn_answer(qbel, sbel1, sbel2, cg1, cg2):
    """
    This method resolves the truthiness of a third order belief question with
    the following form based on the annotations provided.

        Is it the case that {spkr1} believes that {spkr2} believes
        that {spkr1} believes it is {qbel} true that {event}?

    Args:
        qbel (str): question belief (CT-, PS, CT+).
        sbel1 (str): speaker 1 belief (CT-, PS, CT+, NB).
        sbel2 (str): speaker 2 belief (CT-, PS, CT+, NB).
        cg1 (str): speaker 1 common ground (JA, IN, RT, NA).
        cg2 (str): speaker 2 common ground (JA, IN, RT, NA).

    Returns:
        bool: True if the question is correct given the annotations.
    """
    if qbel in ("PS", "CT+") and cg1 in ("JA", "IN") and (
        (qbel == sbel1) or ((qbel == "PS") and (sbel1 == "CT+"))
    ):
        return True
    elif qbel == "CT-" and cg1 in ("RT", "NA") and qbel == sbel1: # NOTE: sbel1 here!!!
        return True
    return False
\end{lstlisting}

\end{document}